\pdfoutput=1

\documentclass[11pt]{article}

\usepackage[final]{acl}

\usepackage{times}
\usepackage{latexsym}

\usepackage[T1]{fontenc}

\usepackage[utf8]{inputenc}

\usepackage{microtype}

\usepackage{inconsolata}

\usepackage{graphicx}

\usepackage{CJKutf8}

\usepackage{algorithm}

\usepackage{algorithmic}

\usepackage{multirow}
\usepackage{makecell}
\usepackage{booktabs}

\usepackage{hyperref}
\usepackage{amsmath}
\usepackage{xcolor}

\usepackage{amsfonts}

\definecolor{mycolor}{RGB}{160, 203, 141}
\definecolor{false_bg}{RGB}{184, 141, 203}
\definecolor{true}{RGB}{50, 160, 90}    
\definecolor{error}{RGB}{220, 60, 50}   
\definecolor{token1}{RGB}{70, 130, 180}   
\definecolor{token2}{RGB}{220, 60, 50}     
\definecolor{token3}{RGB}{50, 160, 90}     
\definecolor{token4}{RGB}{180, 70, 160}    
\definecolor{token5}{RGB}{230, 160, 50}    
\definecolor{token6}{RGB}{0, 150, 150}     

\title{Enhancing Character-Level Understanding in LLMs through Token Internal Structure Learning}

\author{
 \textbf{Zhu Xu}\textsuperscript{$\clubsuit$},
 \textbf{Zhiqiang Zhao}\textsuperscript{$\clubsuit$†},
 \textbf{Zihan Zhang}\textsuperscript{$\clubsuit$},
 \textbf{Yuchi Liu}\textsuperscript{$\clubsuit$},
 \textbf{Quanwei Shen}\textsuperscript{$\clubsuit$},
 \textbf{Fei Liu}\textsuperscript{$\clubsuit$},
\\
 \textbf{Yu Kuang}\textsuperscript{$\clubsuit$},
 \textbf{Jian He}\textsuperscript{$\clubsuit$},
 \textbf{Conglin Liu}\textsuperscript{$\spadesuit$}
\\
\\
\textsuperscript{$\clubsuit$} School of Computer Science and Technology, \\Chongqing University of Posts and Telecommunications
\\
\small{
  s231231076@stu.cqupt.edu.cn, 
  † zhaozq@cqupt.edu.cn,
  s2312310\{91, 46, 51, 42, 31, 20\}@stu.cqupt.edu.cn
}
\\
\textsuperscript{$\spadesuit$} Baidu AI Platform \& Ecosystem
\\
\small{
  liuconglin@baidu.com
}
}

\begin{document}
\maketitle
\begingroup
\renewcommand{\thefootnote}{}
\footnotetext{† Corresponding author.}
\addtocounter{footnote}{-1} 
\begin{abstract}
Tokenization methods like Byte-Pair Encoding (BPE) enhance computational efficiency in large language models (LLMs) but often obscure internal character structures within tokens. This limitation hinders LLMs' ability to predict precise character positions, which is crucial in tasks like Chinese Spelling Correction (CSC) where identifying the positions of misspelled characters accelerates correction processes. We propose \textbf{Token Internal Position Awareness (TIPA)}\footnote[1]{The open-source model and training code can be found at: \url{https://github.com/FloatFrank/TIPA}}, a method that significantly improves models' ability to capture character positions within tokens by training them on reverse character prediction tasks using the tokenizer's vocabulary. Experiments demonstrate that TIPA enhances position prediction accuracy in LLMs, enabling more precise identification of target characters in original text. Furthermore, when applied to downstream tasks that do not require exact position prediction, TIPA still boosts performance in tasks needing character-level information, validating its versatility and effectiveness.
\end{abstract}

\section{Introduction}

Large language models (LLMs) have revolutionized natural language processing by employing tokenization methods such as Byte-Pair Encoding (BPE) \citep{sennrich2015neural, wang2020neural} to segment text into subword units, optimizing computational efficiency. However, BPE often obscures internal character structures within tokens \citep{shin2024large, xu2024llm}, which poses challenges for tasks requiring detailed character-level information.

For instance, LLMs frequently struggle with simple character-counting tasks. When prompted with questions like \textbf{"\textcolor{token1}{How }\textcolor{token2}{many }\textcolor{token3}{r}\textcolor{token4}{'s }\textcolor{token5}{are }\textcolor{token6}{in }\textcolor{token1}{ '}\textcolor{token2}{str}\textcolor{token3}{aw}\textcolor{token4}{berry}\textcolor{token5}{'?}"} \citep{xu2024llm}, many models fail to provide the correct answer due to their limited understanding of character positions within tokens, this limitation is more pronounced in languages like Chinese, where meaning relies heavily on character composition and sequence. Models such as GPT-4o \citep{hurst2024gpt} often misidentify specific character positions in the tokenized text. For example, when asked to locate the character \begin{CJK}{UTF8}{gkai}"\textcolor{token6}{阁}"\end{CJK} in the sentence \begin{CJK}{UTF8}{gkai}"\textcolor{token1}{为什么}\textcolor{token2}{总}\textcolor{token3}{称呼}\textcolor{token4}{对方}\textcolor{token5}{为}\textcolor{token6}{阁}\textcolor{token1}{下}\textcolor{token2}{？}"\end{CJK} (Why do you always address each other as 'Your Excellency'?) \citep{wu2023rethinking}, they frequently provide incorrect positions.

This lack of internal character structure awareness adversely affects LLM performance in character-sensitive applications like Chinese Spelling Correction (CSC), where accurate identification of misspelled characters and their positions is crucial for efficient corrections.

\begin{figure}[t]
  \includegraphics[width=\columnwidth]{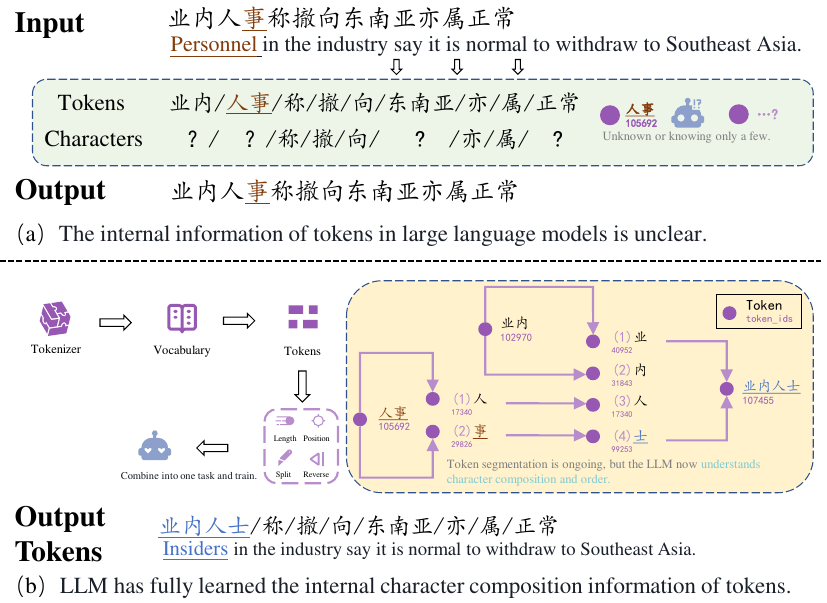}
  \caption{(a) demonstrates LLMs' inability to perform spelling correction correctly without learning token-internal character order. (b) shows TIPA's core dataset construction (left) and its character-level task enhancement through token structure understanding without architectural changes (right).}
  \label{fig:compare_untrain}
\end{figure}

Traditional Transformer-based \citep{vaswani2017attention} language models emphasize next-token prediction, focusing on sequential dependencies between tokens. This focus does not inherently encourage models to capture detailed positional relationships within tokens, leading them to rely more on token order rather than internal character composition.

To address this limitation, we propose \textbf{Token Internal Position Awareness (TIPA)}, a method designed to enhance models' ability to recognize and predict character positions within tokens. TIPA trains LLMs on reverse character prediction tasks using the tokenizer's vocabulary, compelling the model to focus on each character's position independently of sequential context.

Figure~\ref{fig:compare_untrain} illustrates the disparity in CSC performance between untrained and trained LLMs regarding token character composition and sequence.

For example, a token like \begin{CJK}{UTF8}{gkai}"\textcolor{token1}{小说}" (novel)\end{CJK} would be decomposed in TIPA as a JSON structure: \begin{CJK}{UTF8}{gkai}\{2: "\textcolor{token2}{说}", 1: "\textcolor{token3}{小}"\}\end{CJK}, mapping each character to its position in descending order. This approach helps the model develop a structural understanding beyond typical left-to-right reading, which is crucial for tasks requiring precise character positioning.

TIPA leverages the tokenizer’s own vocabulary, allowing the model to internalize character composition and structure without relying on external data, enhancing generalization in position-sensitive tasks.

\begin{table}[t]
\centering
\small
\begin{tabular}{ll}
\toprule
\textbf{Task Type} & \textbf{Example} \\
\midrule
Source & (…106 chars)\begin{CJK}{UTF8}{gkai}网\colorbox{false_bg}{\textcolor{white}{路}}技术有限公司\end{CJK} \\
Traditional Task & (…106 chars)\begin{CJK}{UTF8}{gkai}网\colorbox{mycolor}{\textcolor{white}{络}}技术有限公司\end{CJK} \\
Position Task & [\{108, \begin{CJK}{UTF8}{gkai}\colorbox{false_bg}{\textcolor{white}{路}}, \colorbox{mycolor}{\textcolor{white}{络}}\end{CJK} \}]\\
\bottomrule
\end{tabular}
\caption{Accurately predicting the position of erroneous characters and providing both the incorrect and corrected characters serves two purposes: the incorrect character verifies the model's ability to precisely locate errors, while the corrected character fulfills the error correction task. This approach also reduces the number of output tokens required by the model.}
\label{tab:csc_examples}
\end{table}

Our contributions are:
\begin{enumerate}
    \item \textbf{Enhanced Position Prediction:} Demonstrating the value of accurate character position prediction in CSC tasks, enabling faster and more precise corrections (see Table~\ref{tab:csc_examples}).
    \item \textbf{Introduction of TIPA and MTIPA:} Presenting TIPA and its extension, Multi-Token Internal Position Awareness (MTIPA), which improves models' ability to capture character positions for accurate predictions.
    \item \textbf{Versatility in Downstream Tasks:} Showing that TIPA enhances performance in tasks requiring character-level information, even without explicit position prediction.
\end{enumerate}

\section{Related Work}

Tokenization methods like BPE \citep{sennrich2015neural, wang2020neural} and WordPiece \citep{schuster2012japanese} improve computational efficiency in LLMs but obscure internal character structures. \citet{kaushal2022tokens} found that while larger models encode character-level details better, they may not explicitly understand character positions within tokens. Recent byte-level models like ByT5 \citep{xue2022byt5} process raw bytes for character-level precision but require architectural changes that prevent low-cost adaptation of existing subword-based LLMs. Hybrid approaches (e.g., CANINE \citep{clark2022canine}) improve character awareness but still lack efficient positional modeling for multi-character tokens.

Recent studies \citep{xu2024llm, shin2024large} highlight LLMs' limitations in tasks requiring fine-grained character-level understanding, attributing deficiencies to tokenization and model architecture.

In CSC research, methods like ReLM \citep{liu2024chinese} reframed CSC as sentence rephrasing, while self-supervised learning approaches \citep{jiang2024chinese} showed that models trained on error-free data can outperform those using confusion sets. \citet{li2024c} proposed C-LLM, using character-level tokenization to enhance character-level understanding.

To enhance models' awareness of internal token structures, studies have addressed limitations like the ``reversed curve phenomenon'' \citep{berglund2023reversal, thawani2023learn} and sensitivity to text order \citep{chen2024premise}. \citet{itzhak2022models} found that while models encode orthographic information without direct character-level training, explicitly teaching spelling did not enhance performance. Our work differs by incorporating character position information and reversing character sequences during training, enabling a better understanding of internal token structures and improving tasks like Chinese Spelling Correction.

\section{Methodology}

We introduce two novel techniques: \textbf{Token Internal Position Awareness (TIPA)} and \textbf{Multi-Token Internal Position Awareness (MTIPA)}, designed to enhance large language models' capacity to recognize and leverage internal character structures within tokens.

\subsection{Token Internal Position Awareness (TIPA)}
TIPA leverages the tokenizer's vocabulary to train the model to understand the internal structure of each token. For tokens that can be fully represented in UTF-8 \citep{yergeau2003utf}, we apply a reverse prediction task to capture token-internal positions.

Let $T$ denote the tokenizer, and let $V = \{ t_1, t_2, \dots, t_m \}$ be the set of tokens in the vocabulary of $T$. For each token $t \in V$ that can be fully represented in UTF-8, we decompose $t$ into its constituent characters:
\begin{equation}
C_t = [ c_1, c_2, \dots, c_n ],
\end{equation}
where $n$ is the number of characters in $t$.

We define a reverse position mapping $D_t$ for token $t$ as:
\begin{equation}
D_t = \{ (i, c_{i}) \mid i = n, n-1, \dots, 1 \}.
\end{equation}
This mapping associates each position $i$ (starting from n) with the character at the $i$-th position in $t$, effectively reversing the order of characters. The training prompt template used for this purpose is referenced in Table~\ref{tab:tipa_prompt}.
\begin{algorithm}[H]
\caption{TIPA Algorithm}
\begin{algorithmic}[1]
\REQUIRE Tokenizer $T$
\ENSURE TIPA Dataset $\mathcal{D}_{\text{TIPA}}$
\STATE Initialize $\mathcal{D}_{\text{TIPA}} \leftarrow \emptyset$
\STATE $V \leftarrow$ GetVocabulary($T$)
\FOR{each token $t \in V$}
    \IF{$t$ can be fully represented in UTF-8}
        \STATE Decompose $t$ into characters $C_t = [ c_1, c_2, \dots, c_n ]$
        \STATE Create reverse position mapping $D_t = \{ (i, c_{i}) \mid i = n, n-1, \dots, 1 \}$
        \STATE Add $(t, D_t)$ to $\mathcal{D}_{\text{TIPA}}$
    \ENDIF
\ENDFOR
\STATE Prune irrelevant tokens from $\mathcal{D}_{\text{TIPA}}$
\RETURN $\mathcal{D}_{\text{TIPA}}$
\end{algorithmic}
\end{algorithm}

\textbf{Rationale for Reverse Ordering:} By using reverse ordering, the first number output by the model corresponds to the length of the token ($n$). This approach integrates the token splitting task, length information, and position information into a single method. If we were to use forward ordering (i.e., positions starting from $1$), the model might deduce the length of the token indirectly through the sequence of positions (1, 2, 3, etc.), but it wouldn't inherently know this information. Reverse ordering requires the model to output the token length as the starting position, avoiding this ambiguity.

This method retains the advantages of tokenization while enhancing the model's grasp of character composition and positional information within tokens. It slightly increases the training time due to the additional reverse prediction task but does not introduce any latency during inference.

The TIPA dataset $\mathcal{D}_{\text{TIPA}}$ is then constructed as:
\begin{equation}
\mathcal{D}_{\text{TIPA}} = \{ (t, D_t) \mid t \in V \text{ (UTF-8)} \}.
\end{equation}

An overview of TIPA is illustrated in Figure~\ref{fig:tipa_overview}.

\begin{figure*}[t]
  \includegraphics[width=\linewidth]{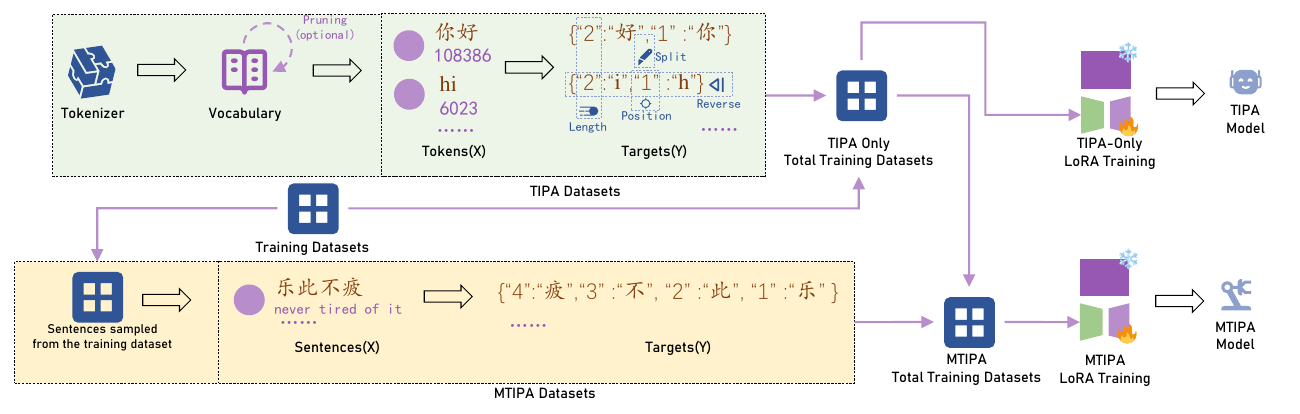}
  \caption{Overview of TIPA and MTIPA. TIPA enhances character-level structure awareness per token. MTIPA extends this to multi-token sequences, enabling fine-grained positional understanding.}
  \label{fig:tipa_overview}
\end{figure*}

\begin{table}[h]
\centering
\small
\begin{tabular}{p{0.9\linewidth}}
\toprule
\textbf{TIPA Prompt Example} \\
\midrule
  \textbf{Instruction}: \begin{CJK}{UTF8}{gkai}直接给出json输出，倒序给出输入的Token中包含的所有位置和字符\end{CJK}\\
  (Translation: Directly output the JSON, listing all characters and their positions in reverse order from the input token.)\\
  \textbf{Input}: girl\\
  \textbf{Output}: \{  "4": "\textbf{l}",  "3": "\textbf{r}",  "2": "\textbf{i}",  "1": "\textbf{g}"\}
\\
\bottomrule
\end{tabular}
\caption{An example of the prompt \citep{zheng2024llamafactory} used in TIPA(←) training, along with its English translation.}
\label{tab:tipa_prompt}
\end{table}

\subsection{Multi-Token Internal Position Awareness (MTIPA)}

Building upon TIPA, we propose \textbf{Multi-Token Internal Position Awareness (MTIPA)} to enhance the model's understanding of character positions within entire sentences or multi-token sequences, especially for tasks that require precise prediction of character positions.

In MTIPA, instead of focusing on individual tokens, we extend the reverse character prediction task to full sentences sampled from the training dataset. This allows the model to learn character positions in the broader context of sentences.

Specifically, we randomly sample a subset of sentences from the target task's training dataset. For each sampled sentence, we decompose it into its constituent characters and create a reverse position mapping, similar to TIPA but applied to the entire sentence.

An overview of MTIPA is illustrated in Figure~\ref{fig:tipa_overview}.

\begin{algorithm}[H]
\caption{MTIPA Algorithm}
\begin{algorithmic}[1]
\REQUIRE Training dataset $\mathcal{D}_{\text{train}}$, sampling ratio $r$
\ENSURE MTIPA dataset $\mathcal{D}_{\text{MTIPA}}$
\STATE Initialize $\mathcal{D}_{\text{MTIPA}} \leftarrow \emptyset$
\STATE Randomly sample a subset $\mathcal{S} \subset \mathcal{D}_{\text{train}}$ with sampling ratio $r$
\FOR{each sentence $s \in \mathcal{S}$}
    \STATE Decompose $s$ into characters $C_s = [ c_1, c_2, \dots, c_n ]$
    \STATE Create reverse position mapping $D_s = \{ (i, c_i) \mid i = n, n-1, \dots, 1 \}$
    \STATE Add $(s, D_s)$ to $\mathcal{D}_{\text{MTIPA}}$
\ENDFOR
\RETURN $\mathcal{D}_{\text{MTIPA}}$
\end{algorithmic}
\end{algorithm}

In practice, we set the sampling ratio $r$ to a small value (e.g., 10\%) to balance the amount of additional data and training efficiency.

MTIPA is specifically applied in tasks that require precise character position prediction within sentences. In our experiments, MTIPA is used in Experiment 1, which involves Chinese Spelling Correction (CSC) with position prediction. In Experiment 2, which focuses on the traditional CSC task without position prediction, MTIPA is not used; only TIPA is applied.  The MTIPA dataset is too long, which can lead to a long training time, and if LoRA training uses a large amount of information to learn how to infer length information, it may reduce the model's ability to perform specific tasks. 

By integrating MTIPA into the training process, the model gains a deeper understanding of character positions in multi-token sequences, leading to improved performance in tasks that demand precise position awareness.

\subsection{Extended Methodology with Full-Parameter SFT}
The tulu-3-sft-mixture dataset\citep{lambert2024tulu3} developed by AI2 enables base LLM refinement into conversational AI. We extracted vocabulary from Llama-3.1\citep{grattafiori2024llama} tokenizer, integrated all tokens into TIPA, and merged this with tulu-3-sft-mixture for full-parameter supervised fine-tuning (SFT) of Llama-3.1-8B. This produced \textbf{Llama-3.1-Tulu-TIPA-8B} without requiring additional character-level task datasets, maintaining the model's inherent capabilities while enhancing character-level processing through the integrated tokenizer vocabulary.

\section{Redefining the Chinese Spelling Correction Task}

\begin{table}[ht]
\centering
\small
\begin{tabular}{lrr}
\hline
\textbf{Dataset} & \textbf{Traditional} & \textbf{Position} \\
\hline
Train & 8,905,800 & 8,016,111 \\
CSCD-Dev & 188,362 & 55,449 \\
CSCD-Test & 188,310 & 54,897 \\
Lemon & 532,684 & 258,112 \\
\hline
\end{tabular}
\caption{Comparison of Output Token Counts between Traditional and Position-based Methods for Various Datasets.}
\label{tab:output_token_comparison}
\end{table}

We redefine the Chinese Spelling Correction (CSC) task to require the model to output the positions of incorrect characters along with their corrections. For example, the sentence \begin{CJK}{UTF8}{gkai}"\textcolor{token6}{业内}\textcolor{token1}{人事}\textcolor{token2}{称}\textcolor{token3}{撤}\textcolor{token4}{向}\textcolor{token5}{东南亚}\textcolor{token6}{亦}\textcolor{token1}{属}\textcolor{token2}{正常}"\end{CJK} (translation: "Industry insiders say that withdrawing to Southeast Asia is also normal") becomes:

\begin{CJK}{UTF8}{gkai}
[{"position": 4, "incorrect": "\textcolor{error}{事}", "correction": "\textcolor{true}{士}"}]
\end{CJK}

Here, the character at position 4, "\begin{CJK}{UTF8}{gkai}\textcolor{error}{事}\end{CJK}", is incorrect and should be corrected to "\begin{CJK}{UTF8}{gkai}\textcolor{true}{士}\end{CJK}".

As shown in Table \ref{tab:output_token_comparison}, the position-based method results in fewer output tokens compared to the traditional method, highlighting its efficiency in this task.

To evaluate model performance under this new framework, we introduce several metrics.

\subsection{Position Prediction Accuracy (PPA)}

Let the source text be $\mathbf{x} = [x_1, x_2, \dots, x_L]$.

When calculating this metric, the characters corrected by the model are not considered, only its ability to perceive locations is evaluated.

Let the model's predicted set of incorrect positions and characters be:

\[
\hat{P} = \{\, (\hat{i}, \hat{c}) \,\}
\]

where $\hat{i}$ is the predicted position, and $\hat{c}$ is the predicted incorrect character at that position.

Position Prediction Accuracy (PPA) is defined as:

\[
\text{PPA} = \frac{ \left|\, \{\, \hat{c} = x_{\hat{i}} \mid (\hat{i}, \hat{c}) \in \hat{P}\,\} \,\right| }{ |\hat{P}| }
\]

That is, PPA measures the proportion of positions predicted by the model where the predicted incorrect character $\hat{c}$ matches the character $x_{\hat{i}}$ at position $\hat{i}$ in the source text $\mathbf{x}$. The denominator $|\hat{P}|$ is the total number of positions predicted by the model.

\subsection{Sentence-Level Accuracy (SA)}

Sentence-Level Accuracy (SA) measures the proportion of sentences where all predicted corrections are entirely accurate, including both the positions and the corrected characters.

\subsection{Sentence-Level Accuracy Ignoring Position (SAIP)}

Sentence-Level Accuracy Ignoring Position (SAIP) calculates the proportion of sentences where all predicted corrections are accurate, regardless of whether the positions were correctly identified.

\subsection{Non-Empty Sample Sentence Accuracy (NESSA)}

Non-Empty Sample Sentence Accuracy (NESSA) evaluates the model's performance on sentences known to contain errors by calculating the proportion of such sentences where all predicted corrections are entirely accurate.

\subsection{Character-Level Precision (CP), Recall (CR), and F1 Score (CF1)}

At the character level \citep{hu2024cscd}:

\begin{itemize}
    \item \textbf{Precision (CP)} is the proportion of incorrect characters identified by the model that is actually incorrect.
    \item \textbf{Recall (CR)} is the proportion of actual incorrect characters that are correctly identified by the model.
    \item \textbf{F1 Score (CF1)} is the harmonic mean of Precision and Recall.
\end{itemize}

\section{Experiments}

We conducted comprehensive experiments to validate the effectiveness of TIPA and MTIPA. Our experiments are divided into three main parts: position-aware CSC, traditional CSC, and general model training and evaluation. Below, we detail the datasets, baselines, and training protocols.

\subsection{Datasets}

Following the method of C-LLM \citep{li2024c, li2022improving, liang2023disentangled}, we selected two new Chinese Spelling Correction (CSC) benchmarks, CSCD-NS and LEMON, to address the limitations identified \citep{hu2024cscd, yin2023comprehensive, li2022improving} in previous datasets like SIGHAN \citep{wu2013chinese, yu2014chinese, tseng2015introduction, sun2024two}. Additionally, we incorporated a large-scale pseudo-data set, Wang271K, generated by ASR or OCR methods, to enhance our training set. The datasets used for training and evaluation are as follows:

\begin{itemize}
\item \textbf{Wang271K} \citep{wang2018hybrid}: A dataset containing 271,329 sentences with errors introduced based on linguistic rules. This dataset was used in combination with CSCD-NS for training.
\item \textbf{CSCD-NS} \citep{hu2024cscd}: A high-quality CSC dataset where the primary source of character errors stems from pinyin input methods. It contains a significant amount of homophonic and word-level errors, making it superior to SIGHAN. The validation data from CSCD-NS was used as our validation set.
\item \textbf{LEMON} \citep{wu2023rethinking}: A novel, large-scale, multi-domain CSC dataset featuring various real-world spelling errors. It spans seven different sub-domains, including game (GAM), encyclopedia (ENC), contract (COT), medical care (MEC), car (CAR), novel (NOV), and news (NEW), typically testing the model’s domain correction capabilities in a zero-shot setting.
\end{itemize}

The training set was composed of the combined data from CSCD-NS and Wang271K. The validation set was derived from CSCD-NS, and the models were tested on both the CSCD-NS test data and LEMON.

For \textbf{Experiment 1}, we combined the training sets of Wang271K and CSCD-NS as our training data. The validation set was the CSCD-NS validation set. We tested our models on the CSCD-NS test set. \citep{li2024c}

For \textbf{Experiment 2}, we followed the same data split but focused on the traditional CSC task without explicit position prediction. In addition, we also tested the model on the Lemon dataset.

For \textbf{general model} training, we used \textbf{tulu-3-sft-mixture}, an open-source conversational AI fine-tuning dataset for the supervised fine-tuning (SFT) stage\citep{lambert2024tulu3}. We performed full-parameter supervised fine-tuning of Llama-3.1-8B using the combined tulu-3-sft-mixture and TIPA dataset. The TIPA dataset was constructed by extracting vocabulary from the Llama-3.1 tokenizer and integrating all tokens into the TIPA format.

For general model evaluation, we used the following benchmarks: \textbf{IFEVAL}\citep{zhou2023instruction} (instruction following evaluation with strict/loose scoring), \textbf{GSM8K} \citep{cobbe2021training} (math word problem-solving), \textbf{MMLU}\citep{hendrycks2020measuring} (massive multitask language understanding), \textbf{AExams}\citep{hardalov2020exams} (Arabic exam question answering), \textbf{KoBEST}\citep{jang2022kobest} (Korean language understanding benchmark), and \textbf{HumanEval}\citep{chen2021evaluating} (Python code generation), \textbf{TyDi QA}\citep{clark2020tydi} (Multilingual QA benchmark). Inspired by LLM The Genius Paradox\citep{xu2024llm}, we developed multilingual datasets covering three tasks: Occurrence, Length, and Distinct.

\subsection{Baseline Models}

We compared our methods against several baseline models, including Pure-SFT, GPT-4o \citep{hurst2024gpt}, DeepSeek v2.5, ERNIE-4.0, and GLM-4-Plus, to assess the relative performance improvements. For general model evaluation, we used Llama-3.1-Tulu as a baseline.

\subsection{Training Details}

In Experiment 1 and Experiment 2, we fine-tuned the open-source Qwen2.5-7B \citep{yang2024qwen2} model using LoRA \citep{hu2021lora} to incorporate TIPA and MTIPA. LoRA allows efficient adaptation of large models without modifying the original model weights, making it suitable for our experiments.

For TIPA integration, we generated a TIPA dataset by performing set deduplication on tokens appearing in Wang271K, CSCD-NS, and LEMON to reduce the number of tokens for TIPA operations. This does not mean that preference learning has been applied to the dataset, because the result of pruning more than 300,000 pieces of data is only 24,994 tokens, which is sufficient to include all Chinese character tokens. This can be considered as a preference for Chinese characters, which is also a learning strategy. In practical applications, we can directly perform TIPA on all tokens of the tokenizer.

For MTIPA integration, we randomly sampled 10\% of the training set and constructed the MTIPA dataset using the same method as TIPA but applied to multi-token sequences.

In Experiment 3, we extracted all tokens from the Llama-3.1 tokenizer vocabulary, performed TIPA operations to generate a dataset, and conducted mixed training with the tulu-3-sft-mixture dataset. Additional training details, hardware configurations, and training time can be found in Appendix \ref{subsec:implementation}.

\subsection{Experiment 1: CSC with Position Prediction}

In this experiment, we evaluated the models on the redefined CSC task requiring position prediction.

\subsubsection{Results}

\begin{table*}[ht]
  \centering
  \small
  \begin{tabular}{l|ccccccc}
    \toprule
    Model & PPA (\%) & SA (\%) & SAIP (\%) & NESSA (\%) & CP (\%) & CR (\%) & CF1 (\%)\\
    \midrule
Qwen2.5-7B & 4.03 & 32.76 & 37.62 & 0.61 & 0.86 & 1.42 & 1.07 \\
GPT-4o & 11.14 & 43.76 & 48.68 & 2.56 & 3.20 & 4.31 & 3.68 \\
DeepSeek v2.5 & 6.67 & 49.60 & 50.88 & 0.65 & 0.63 & 1.98 & 0.95 \\
GLM-4-Plus & 13.53 & 38.52 & 43.46 & 4.39 & 1.18 & 7.00 & 2.02 \\
ERNIE-4.0 & 4.06 & 40.10 & 43.07 & 0.72 & 1.60 & 3.64 & 2.22 \\
    \midrule
Pure-SFT-7B & 79.45 & 69.58 & 74.64 & 49.33 & 56.61 & 50.12 & 53.17 \\
TIPA-7B & $84.72\textsuperscript{↑5.27}$ & $70.70\textsuperscript{↑1.12}$ & $75.90\textsuperscript{↑1.26}$ & $51.63\textsuperscript{↑2.30}$ & $58.72\textsuperscript{↑2.11}$ & $51.54\textsuperscript{↑1.42}$ & $54.90\textsuperscript{↑1.73}$ \\
MTIPA-7B\textsubscript{($r$=10\%)} & $\textbf{87.52}\textsuperscript{↑8.07}$ & $\textbf{72.40}\textsuperscript{↑2.82}$ & $\textbf{77.00}\textsuperscript{↑2.36}$ & $\textbf{54.67}\textsuperscript{↑5.34}$ & $\textbf{63.25}\textsuperscript{↑6.64}$ & $\textbf{54.95}\textsuperscript{↑4.83}$ & $\textbf{58.81}\textsuperscript{↑5.64}$ \\
\bottomrule
  \end{tabular}
  \caption{\label{Experiment 1 Results}(Experiment 1)Results on position-based CSC using CSCD-NS test dataset, showing that TIPA-7B and MTIPA-7B outperform the baseline in all evaluated metrics. GPT-4o's advantage stems from its character-level tokenization simplifying position identification, while other models face challenges with subword tokenization. Cross-tokenizer comparisons are less meaningful, yet previous LLMs performed poorly on this task.}
\end{table*}

Table~\ref{Experiment 1 Results} presents the performance comparison. Our TIPA-7B model outperformed the baseline Pure-SFT-7B across all metrics, demonstrating the effectiveness of TIPA in enhancing position awareness. The MTIPA-7B model further improved performance, indicating that incorporating multi-token sequences benefits the model's understanding of character positions.

\subsection{Experiment 2: Traditional CSC Task}

In the second experiment, we assessed the models on the traditional CSC task, which does not involve explicit position prediction. We also compared the impact of using forward (→) and reverse (←) TIPA constructions. This task is based on single-character to single-character mappings, and the evaluation methods are similarly structured. Due to the inherent difficulty large models like GPT-4o face in producing outputs with equal character lengths, we excluded any non-equal length data from the evaluation metrics. This exclusion ensures that length inconsistencies do not skew the experimental results. Detailed information on the models' output length consistency is available in the appendix. Some models prefer to convert half-width symbols to full-width ones, and we treat them as correct by establishing a mapping between half-width and full-width characters.

\subsubsection{Results}

Table~\ref{exp-results-traditional} shows that TIPA improves model performance even when position prediction is not required. The reverse TIPA construction (←) consistently outperforms the forward version (→), suggesting that reverse ordering better enhances the model's understanding of internal character structures.

\begin{table*}[t]
  \centering
  \small
  \begin{tabular}{l|ccc|ccc|ccc|ccc}
    \toprule
    \multirow{3}{*}{\makecell{Model}} & \multicolumn{6}{c|}{Sentence Level} & \multicolumn{6}{c}{Character Level}\\
    \cline{2-13}
    & \multicolumn{3}{c|}{Detection} & \multicolumn{3}{c|}{Correction} & \multicolumn{3}{c|}{Detection} & \multicolumn{3}{c}{Correction} \\
    \cline{2-13}
    & P & R & F1 & P & R & F1 & P & R & F1 & P & R & F1\\
\midrule
Qwen2.5-7B & 45.69 & 57.99 & 51.11 & 42.16 & 53.51 & 47.16 & 31.50 & 68.11 & 43.08 & 27.44 & 59.33 & 37.52 \\
GPT-4o & 35.88 & 65.19 & 46.28 & 33.36 & 60.62 & 43.03 & 34.10 & 86.64 & 48.94 & 31.14 & 79.10 & 44.68 \\
\midrule
Pure-SFT-1.5B & 50.06 & 43.59 & 46.60 & 42.89 & 37.34 & 39.92 & 53.05 & 49.32 & 51.12 & 42.93 & 39.91 & 41.36 \\
TIPA(→)-1.5B & \textbf{75.58} & 62.11 & 68.18 & \textbf{71.55} & \textbf{58.80} & \textbf{64.55} & \textbf{77.27} & 64.38 & \textbf{70.24} & \textbf{71.91} & 59.91 & \textbf{65.36}\\
TIPA(←)-1.5B & 75.43 & \textbf{62.21} & \textbf{68.19} & 71.19 & 58.72 & 64.36 & 75.60 & \textbf{64.58} & 69.66 & 70.21 & \textbf{59.98} & 64.69  \\
\midrule
Pure-SFT-3B & 78.39 & 66.96 & 72.22 & 74.87 & 63.96 & 68.99 & 80.59 & 69.25 & 74.49 & 75.54 & 64.91 & 69.82 \\
TIPA(→)-3B & 77.80 & 67.73 & 72.42 & 75.10 & 65.38 & 69.91 & 79.49 & 69.80 & 74.33 & 75.63 & 66.41 & 70.72 \\
TIPA(←)-3B & \textbf{78.78} & \textbf{68.47} & \textbf{73.26} & \textbf{75.83} & \textbf{65.90} & \textbf{70.52} & \textbf{80.88} & \textbf{70.88} & \textbf{75.55} & \textbf{76.52} & \textbf{67.06} & \textbf{71.47} \\
\midrule
Pure-SFT-7B & 78.46 & 69.07 & 73.47 & 75.44 & 66.42 & 70.64 & 80.73 & 70.94 & 75.52 & 77.05 & 67.70 & 72.07 \\
TIPA(→)-7B & 78.04 & 69.56 & 73.55 & 75.32 & 67.14 & 70.99 & 79.98 & 72.63 & 76.13 & 76.76 & 69.70 & 73.06 \\
TIPA(←)-7B & \textbf{81.79} & \textbf{70.95} & \textbf{75.98} & \textbf{78.84} & \textbf{68.40} & \textbf{73.25} & \textbf{83.33} & \textbf{73.34} & \textbf{78.01} & \textbf{79.64} & \textbf{70.09} & \textbf{74.56} \\
\bottomrule
  \end{tabular}
  \caption{\label{exp-results-traditional}(Experiment 2)Traditional CSC results on CSCD-NS test dataset. TIPA improves character-level detection and correction, with reverse-order TIPA showing greater gains at larger model scales.}
\end{table*}

\subsubsection{Evaluation on LEMON Dataset}

To further assess the generalization ability of our models, we evaluated them on the LEMON dataset, which contains longer and more complex sentences. Table~\ref{exp-results-lemon} shows that our TIPA-7B model achieves higher character-level F1 scores across various subsets of the LEMON dataset, indicating improved performance on difficult cases.

\begin{table*}[t]
  \centering
  \small
  \begin{tabular}{l|ccccccccc}
    \toprule
    \multirow{2}{*}{Model} & \multicolumn{9}{c}{Character-Level F1 Score (\%)} \\
    \cline{2-10}
    & CAR & COT & ENC & GAM & MEC & NEW & NOV & CSCD-NS & AVG\\
\midrule
Qwen2.5-7B & 34.04 & 50.35 & 46.73 & 25.34 & 53.51 & 34.3 & 28.99 & 37.52 & 38.85 \\
GPT-4o & 41.87 & 44.11 & 47.98 & 31.62 & 51.48 & 47.11 & 37.52 & 44.68 & 43.3 \\
\midrule
Pure-SFT-1.5B & 44.2 & 52.73 & 44.11 & 27.95 & \textbf{51.21} & \textbf{48.59} & 29.31 & 41.36 & 42.43 \\
TIPA(→)-1.5B & \textbf{45.20} & \textbf{52.95} & \textbf{46.19} & 28.4 & 50.03 & 47.41 & \textbf{29.68} & \textbf{65.36} & \textbf{45.65} \\
TIPA(←)-1.5B & 44.92 & 50.81 & 44.69 & \textbf{28.49} & 50.81 & 48.46 & 29.09 & 64.69 & 45.24 \\
\midrule
Pure-SFT-3B & 49.18 & 59.34 & 46.93 & 26.13 & 55.15 & \textbf{56.25} & 32.68 & 69.82 & 49.44 \\
TIPA(→)-3B & \textbf{49.72} & 59.47 & 48.54 & 32.71 & 55.12 & 55.44 & 32.85 & 70.72 & 50.57 \\
TIPA(←)-3B & 49.22 & \textbf{60.51} & \textbf{48.56} & \textbf{34.67} & \textbf{56.02} & 55.8 & \textbf{33.32} & \textbf{71.47} & \textbf{51.20} \\
\midrule
Pure-SFT-7B & 52.47 & 58.77 & 53.28 & 32.32 & 61.62 & 60.27 & 35.41 & 72.07 & 53.28 \\
TIPA(→)-7B & \textbf{56.07} & \textbf{64.02} & 52.91 & 35.56 & 62.56 & 60.24 & 38.96 & 73.06 & 55.42 \\
TIPA(←)-7B & 53.69 & 59.79 & \textbf{55.65} & \textbf{36.46} & \textbf{63.15} & \textbf{61.16} & \textbf{39.65} & \textbf{74.56} & \textbf{55.51} \\
\bottomrule
  \end{tabular}
  \caption{\label{exp-results-lemon}(Experiment 2)Character-level F1 scores on LEMON and CSCD-NS for various domains. TIPA-7B consistently achieves higher F1, indicating better generalization.}
\end{table*}

\subsection{Experiment 3: General Model Evaluation}

To validate TIPA's effectiveness beyond Chinese-specific tasks and LoRA fine-tuning, we conducted full-parameter supervised fine-tuning experiments on the Llama-3.1-8B model.

\subsubsection{Results}

Table~\ref{tab:standard_benchmarks} shows that TIPA-enhanced models maintain comparable performance on benchmarks while improving in character-sensitive tasks. The TIPA model achieves particularly strong gains in IFEVAL (strict and loose) and AExams, demonstrating enhanced instruction following capabilities.

\begin{table}[t]
  \centering
  \small
  \begin{tabular}{lcc}
  \toprule
  \textbf{Metric} & \textbf{TULU} & \textbf{TULU-TIPA} \\
  \midrule
  IFEVAL\textsubscript{0-shot, strict↑} & 64.88 & \textbf{67.84} \\
  IFEVAL\textsubscript{0-shot, loose↑} & 68.02 & \textbf{70.24} \\
  GSM8K\textsubscript{8-shot↑} & \textbf{74.53} & \textbf{74.53} \\
  MMLU\textsubscript{5-shot↑} & \textbf{65.30} & \textbf{65.30} \\
  AExams\textsubscript{↑} & 38.92 & \textbf{40.22} \\
  KoBEST\textsubscript{F1↑} & 51.09 & \textbf{52.37} \\
  HumanEval\textsubscript{pass@1↑} & \textbf{53.05} & 51.83 \\
  \bottomrule
  \end{tabular}
  \caption{Standard benchmark evaluation of \textbf{Llama-3.1-Tulu} and \textbf{Llama-3.1-Tulu-TIPA-8B}, showing comparable performance with TIPA-enhanced model.}
  \label{tab:standard_benchmarks}
\end{table}

Table~\ref{tab:tydi_qa} demonstrates TIPA's effectiveness in multilingual settings, with the average F1 score improving from 47.85 to 52.81 across nine languages. The most significant gains appear in Finnish, Indonesian, and Korean, suggesting TIPA helps with non-Latin scripts.

\begin{table}[t]
  \centering
  \small
  \begin{tabular}{lcccccc}
  \toprule
  \textbf{Language} & \multicolumn{2}{c}{\textbf{F1}} \\
   & TULU & TULU-TIPA \\
  \midrule
  Arabic & \textbf{67.22} & 67.02 \\
  Bengali &  30.44 & \textbf{31.13} \\
  English &  61.16 & \textbf{64.83} \\
  Finnish & 49.74 & \textbf{60.99} \\
  Indonesian& 56.76 & \textbf{66.64} \\
  Korean & 46.16 & \textbf{56.37} \\
  Russian &  44.07 &\textbf{44.95} \\
  Swahili & 47.73 & \textbf{53.10} \\
  Telugu & 27.39 & \textbf{30.28} \\
  \hline
  \textbf{Average} &  47.85 & \textbf{52.81} \\
  \bottomrule
  \end{tabular}
  \caption{TyDi QA multilingual evaluation showing improved performance with TIPA.}
  \label{tab:tydi_qa}
\end{table}

Table~\ref{tab:char_avg} confirms TIPA's benefits for character-level tasks, with the TIPA model outperforming baseline by 3.95\% on character occurrence counting, 7.11\% on sentence length prediction, and achieving double the performance on distinct character counting (9.29\% vs 18.74\%).

\begin{table}[t]
  \centering
  \small
  \begin{tabular}{lcc}
  \toprule
  \textbf{Metric} & \textbf{TULU} & \textbf{TULU-TIPA} \\
  \midrule
  Occurrence      & 34.62         & \textbf{38.57} \\
  Length          & 24.48         & \textbf{31.59} \\
  Distinct        & 9.29          & \textbf{18.74} \\
  \bottomrule
  \end{tabular}
  \caption{Average character-level comparison between TULU and TIPA systems across eight languages (CN, EN, JP, KR, AR, EN-Hard, FR, RU).}
  \label{tab:char_avg}
\end{table}

\section{Analysis}

We conducted an in-depth analysis to understand the impact of TIPA and MTIPA on model performance.

\subsection{Position Prediction Accuracy}

Figure~\ref{fig:position_abilities} compares position prediction accuracy across different character positions. The MTIPA-7B model consistently outperforms others, especially at higher character positions, indicating its enhanced ability to handle longer sequences.

\begin{figure}[t]
  \includegraphics[width=\columnwidth]{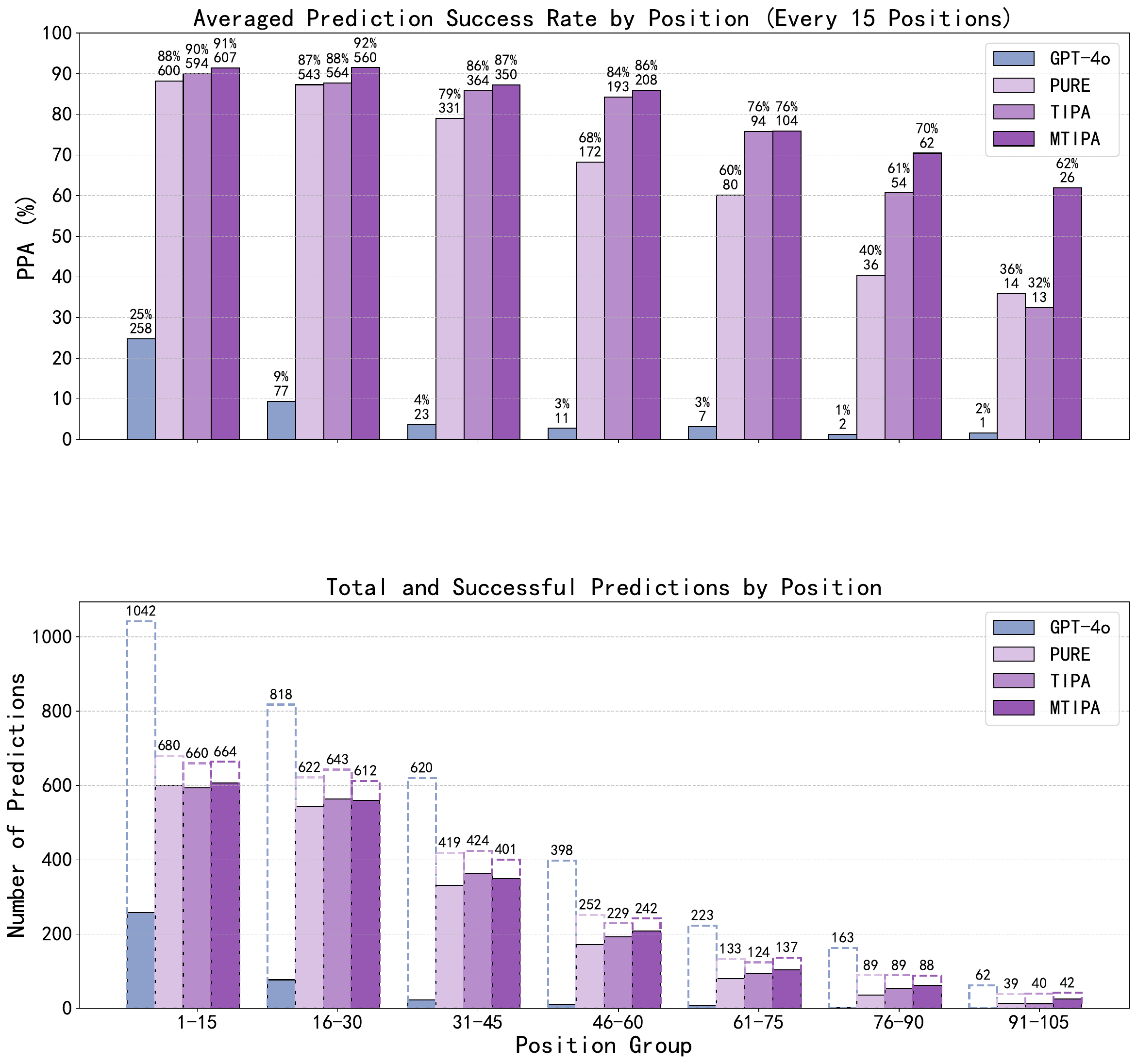}
  \caption{(Experiment 1)Comparison of position prediction accuracy by character position. MTIPA-7B achieves consistently higher accuracy, especially at longer positions.}
  \label{fig:position_abilities}
\end{figure}

\subsection{Training Dynamics}

Figure~\ref{fig:metrics_comparison} shows the comparison of character-level metrics and position accuracy across different epochs for Pure-SFT-7B, TIPA-7B, and MTIPA-7B. MTIPA-7B achieves the highest performance ceiling.

\begin{figure}[t]
  \includegraphics[width=\columnwidth]{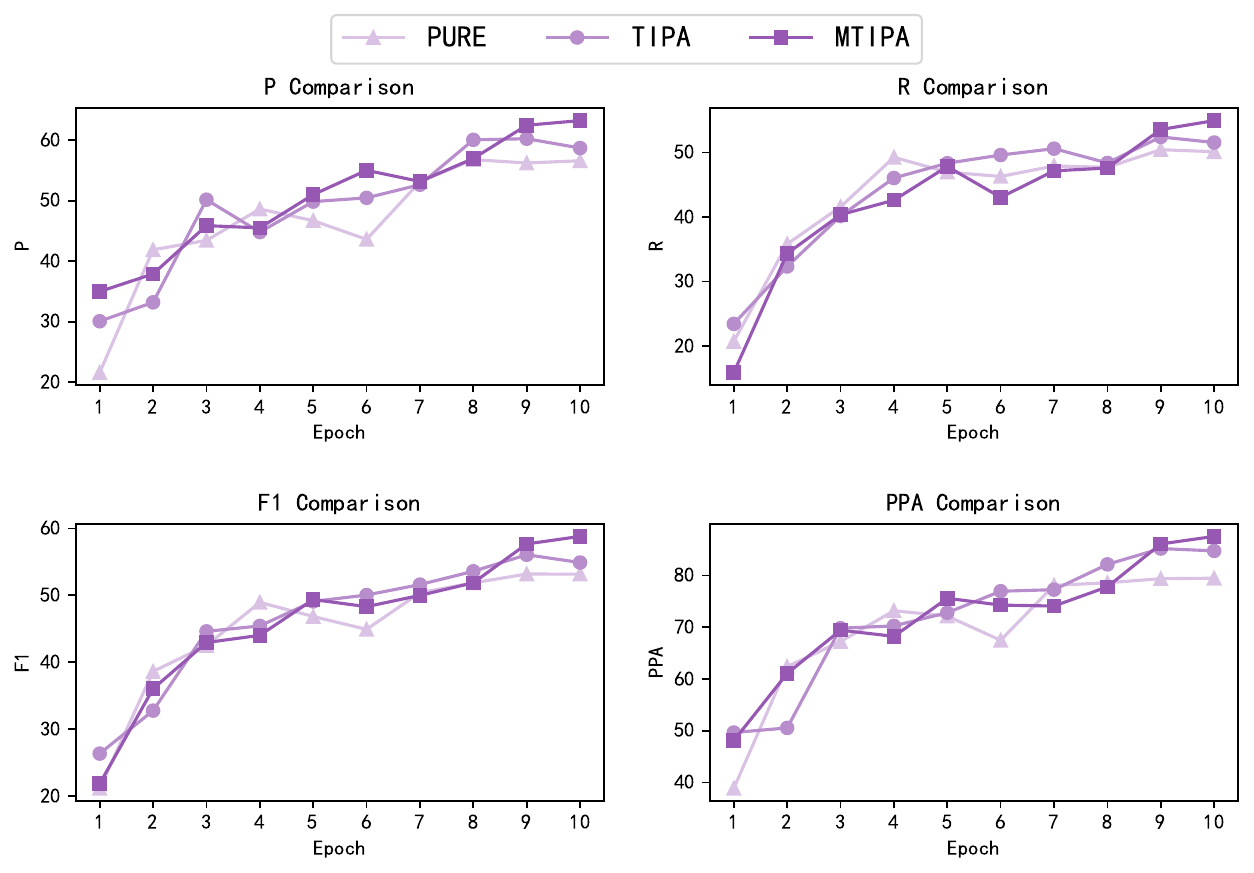}
  \caption{(Experiment 1)Training dynamics showing character-level metrics and position accuracy. MTIPA-7B achieves the highest performance ceiling.}
  \label{fig:metrics_comparison}
\end{figure}

\subsection{Impact on Downstream Tasks}

Our analysis confirms that TIPA and MTIPA significantly enhance models' understanding of internal character structures, leading to improved performance in both position prediction and traditional CSC tasks. The models no longer need to ``guess'' the composition of tokens, as the training includes explicit character structure information.

Furthermore, by not altering the tokenizer or model architecture, our methods maintain compatibility with existing systems and do not introduce additional inference latency.

\subsection{General Model Findings}

Our general model experiments reveal several key insights:
\begin{itemize}
\item TIPA integration through full-parameter SFT maintains model performance on benchmarks while improving character-level capabilities
\item The multilingual evaluation shows TIPA's benefits extend beyond Chinese to other scripts and languages
\item Character-level tasks demonstrate significant improvements, particularly in counting distinct characters
\item The approach scales effectively to larger models (8B parameters) without performance degradation
\end{itemize}

These results suggest that TIPA's benefits are not limited to Chinese-specific tasks or LoRA fine-tuning, but represent a general improvement in LLMs' character-level understanding.

\section{Conclusion}

We introduced \textbf{Token Internal Position Awareness (TIPA)} and \textbf{Multi-Token Internal Position Awareness (MTIPA)} to enhance large language models' (LLMs) ability to accurately predict character positions within tokens. Our experiments demonstrate TIPA's effectiveness across three scenarios: position-aware Chinese spelling correction, traditional CSC tasks, and both training and evaluation of general models. The method shows particular strength in:

1. Improving position prediction accuracy in Chinese text (up to 8.07\% absolute gain)
2. Enhancing traditional CSC performance (up to 5.64\% F1 improvement)
3. Boosting multilingual and character-level capabilities without sacrificing benchmark performance
4. Scaling effectively to larger models through both LoRA and full-parameter fine-tuning

By training LLMs on reverse character prediction tasks using the tokenizer's vocabulary, TIPA, and MTIPA effectively address the limitations imposed by tokenization methods like Byte-Pair Encoding (BPE) while maintaining compatibility with existing systems.

\section{Limitations}

While our study presents promising results, there are several limitations:

\begin{itemize}
  \item The methods' ability to predict positions for out-of-vocabulary (OOV)\citep{sennrich2015neural, wang2020neural} words, which do not directly appear in the tokenizer's vocabulary, requires further investigation.
  \item Directly mixing TIPA datasets during general model SFT training may induce a bias toward shorter text sequence generation when using excessive data proportions. While this can be mitigated by setting minimum token constraints, more fundamental solutions through pretraining or reinforcement learning stages warrant further exploration.
\end{itemize}

\section{Acknowledgments}
This work was supported by the "Major Special Project for Technological Innovation and Application Development" (Grant No. CSTB2024TIAD-STX0036) from the Chongqing Municipal Science and Technology Bureau. We sincerely appreciate the insightful comments and constructive suggestions from the reviewers, which significantly improved the quality and rigor of this manuscript. We are especially grateful to Dr. Yixian Shen, a postdoctoral researcher at the University of Amsterdam, for generously providing computational resources and invaluable guidance during the large-scale model training phase. Their expertise and support were instrumental in advancing this research.

\bibliography{custom}

\appendix

\section{Appendix}
\label{sec:appendix}

\subsection{The limitations of GPT-4o}
In Table~\ref{tab:gpt4o}, we present examples highlighting the limitations of GPT-4o in handling character-level tasks due to its lack of internal character structure awareness.

\begin{table*}[ht]
\centering
\small
\begin{tabular}{p{15cm}}
\toprule
\textbf{Example 1} \\
\midrule
\textbf{Prompt:} \begin{CJK}{UTF8}{gkai}这句话中有几个字符: 我今天没有吃午饭非常饿\end{CJK} \\
Translation: How many characters are in this sentence: I skipped lunch today and am now ravenous.

\textbf{GPT-4o Response:}

\begin{CJK}{UTF8}{gkai}这句话中共有 13个字符，分别是：

我 今天 没 有 吃 午 饭 非 常 饿\end{CJK}

\textbf{Analysis:}  GPT-4o incorrectly states that there are 13 characters and fails to separate \begin{CJK}{UTF8}{gkai}“今天”\end{CJK} into two characters. The correct character count is 11, and \begin{CJK}{UTF8}{gkai}“今天”\end{CJK} should be split into \begin{CJK}{UTF8}{gkai}“今”\end{CJK} and \begin{CJK}{UTF8}{gkai}“天”\end{CJK}.

\\
\midrule

\textbf{Example 2} \\
\midrule

\textbf{Prompt:} \begin{CJK}{UTF8}{gkai}拆分成字符倒序输出金庸武侠小说，直接给出\end{CJK} \\
Translation: Split into characters and output in reverse order "Jin Yong Wuxia novel", output directly.

\textbf{GPT-4o Response:} \begin{CJK}{UTF8}{gkai}小说侠武庸金\end{CJK}

\textbf{Analysis:} GPT-4o incorrectly treats \begin{CJK}{UTF8}{gkai}“小说”\end{CJK} ("novel") as a single character, whereas it comprises two characters \begin{CJK}{UTF8}{gkai}“小”\end{CJK} and \begin{CJK}{UTF8}{gkai}“说”\end{CJK}. The correct reverse order should be \begin{CJK}{UTF8}{gkai}“说小侠武庸金”\end{CJK}. When asked to separate characters with commas, GPT-4o responds:

\begin{CJK}{UTF8}{gkai}小说，侠武，庸金\end{CJK}

Again, it incorrectly groups the characters, demonstrating a lack of internal character awareness.

\\
\midrule

\textbf{Example 3} \\
\midrule

\textbf{Prompt:} \begin{CJK}{UTF8}{gkai}纠正输入这段话中的错别字，以[\{position: 字符位置, incorrect: 错误字符, correction: 纠正后的字符\}, ...]形式给出，字符位置从1开始计数，如果全部正确，给出[]：

并计划在几年底发展到15各分公司，明年增加到20个分公司\end{CJK} \\
Translation: Correct the typos in this sentence, and present in the form of [\{position: character position, incorrect: incorrect character, correction: corrected character\}, ...], counting character positions starting from 1. If all is correct, output []:

And plan to develop 15 branch companies by the end of the year, and increase to 20 branch companies next year.

\textbf{GPT-4o Response:}

\begin{CJK}{UTF8}{gkai}[
    \{"position": 6, "incorrect": "各", "correction": "个"\}
]\end{CJK}

\textbf{Analysis:} GPT-4o correctly identifies the misspelled character \begin{CJK}{UTF8}{gkai}“各”\end{CJK} and suggests the correction \begin{CJK}{UTF8}{gkai}“个”\end{CJK}, but it incorrectly reports the position as 6. The actual position of the error in the sentence is 13, indicating a misunderstanding of character positions within the tokenized text.
\\
\bottomrule
\end{tabular}
\caption{Examples demonstrating GPT-4o's limitations in character-level tasks due to lack of internal character structure awareness.}
\label{tab:gpt4o}
\end{table*}

These examples illustrate how GPT-4o struggles with tasks that require precise character-level understanding. The tokenization process obscures internal character structures, leading to errors in character counting, splitting, and position identification. Our proposed methods, TIPA and MTIPA, aim to address these issues by enhancing models' awareness of internal token structures.

\subsection{Implementation Details}
\label{subsec:implementation} 

In our experiments, we utilized PyTorch and the Hugging Face Transformers \citep{wolf2020transformers} library to implement our models. We fine-tuned the open-source models using LoRA \citep{hu2021lora} (Low-Rank Adaptation of Large Language Models) with specific configurations to efficiently adapt the large models without modifying the original weights.

\subsubsection{TIPA Training Methodology}
We implemented two approaches for TIPA training:

\begin{itemize}
    \item \textbf{Unpruned TIPA}: We first extracted all vocabulary from the tokenizer and filtered out tokens that could not be properly parsed as UTF-8. We then applied the TIPA method to create the TIPA dataset, which was mixed with the CSC dataset and shuffled before LoRA fine-tuning.
    
    \item \textbf{Pruned TIPA}: We tokenized a large CSC dataset and intersected these tokens with the tokenizer's vocabulary. This pruned approach resulted in a TIPA dataset containing primarily Chinese characters and a small portion of other language tokens, while still covering nearly all Chinese characters. This dataset was then mixed with the CSC dataset and shuffled before LoRA fine-tuning.
\end{itemize}

\subsubsection{MTIPA Training Methodology}
For MTIPA implementation, we:

\begin{itemize}
    \item Selected either the unpruned or pruned TIPA method to create the base TIPA dataset
    \item Randomly sampled incorrect source texts from the dataset
    \item Decomposed the sampled strings using the same reverse decomposition task as for individual tokens
    \item Mixed the resulting MTIPA dataset with the CSC and TIPA datasets
    \item Shuffled the combined dataset before LoRA fine-tuning
\end{itemize}

This combined training approach ensures that the model learns both token-internal structures and their application in context while maintaining the efficiency of standard fine-tuning procedures.

For the TIPA and MTIPA methods, we applied LoRA \citep{hu2021lora} with a rank of 16, an alpha of 16, and a dropout rate of 0.05. The optimizer used was AdamW. The training was conducted on a single NVIDIA A800 GPU with 80GB of memory.

The configurations for Experiment 1 are detailed in Table~\ref{tab:config_exp1}, and the training speed and resource consumption are summarized in Table~\ref{tab:train_speed_exp1}.

\begin{table}[h]
  \centering
   \small
  \begin{tabular}{lc}
    \toprule
    \textbf{Configs} & \textbf{Values} \\
    \midrule
    Devices & 1 NVIDIA A800 GPU (80GB) \\
    Batch size     & 16 \\
    Learning rate     & $1 \times 10^{-4}$ \\
    Epochs     & 10 \\    
    LoRA rank     & 16 \\    
    LoRA alpha     & 16 \\            
    LoRA dropout     & 0.05 \\                
    Optimizer & AdamW \\
    \bottomrule
  \end{tabular}
  \caption{Configurations for Experiment 1.}
  \label{tab:config_exp1}
\end{table}

\begin{table}[h]
  \centering
  \small
  \begin{tabular}{lccc}
    \toprule
    \textbf{Method} &  \textbf{Batches}  & \textbf{Speed} & \textbf{GPU hours} \\
    \midrule
    Pure-SFT-7B & 188,340 &  $\sim1$ s/batch  & 52.6 h\\
    TIPA-7B & 203,960 &  $\sim1$ s/batch &  56.7 h\\
    MTIPA-7B & 222,780 &  $\sim1.6$ s/batch & 99.0 h\\ 
    \bottomrule
  \end{tabular}
  \caption{Training speed and resource consumption for Experiment 1.}
  \label{tab:train_speed_exp1}
\end{table}

Similarly, the configurations for Experiment 2 are presented in Table~\ref{tab:config_exp2}, and the training speed is shown in Table~\ref{tab:train_speed_exp2}.

\begin{table}[h]
  \centering
  \small
  \begin{tabular}{lc}
    \toprule
    \textbf{Configs} & \textbf{Values} \\
    \midrule
    Devices & 1 NVIDIA A800 GPU (80GB) \\
    Batch size     & 16 \\
    Learning rate     & $1 \times 10^{-4}$ \\
    Epochs     & 6 (results reported at epoch 3) \\    
    LoRA rank     & 16 \\    
    LoRA alpha     & 16 \\            
    LoRA dropout     & 0.05 \\                
    Optimizer & AdamW \\
    \bottomrule
  \end{tabular}
  \caption{Configurations for Experiment 2. The models achieved optimal performance at epoch 3, beyond which overfitting was observed.}
  \label{tab:config_exp2}
\end{table}

\begin{table}[h]
  \centering
    \small
  \begin{tabular}{lccc}
    \toprule
    \textbf{Method} &  \textbf{Batches}  & \textbf{Speed} & \textbf{GPU hours} \\
    \midrule
    Pure-SFT-7B & 56,502 & $\sim0.96$ s/batch  & $\sim15$ h\\
    TIPA(→)-7B & 61,188 & $\sim0.95$ s/batch  & $\sim16$ h\\ 
    TIPA(←)-7B & 61,188 & $\sim0.95$ s/batch  & $\sim16$ h\\ 
    Pure-SFT-3B & 56,502 &  $\sim0.58$ s/batch & $\sim9$ h\\
    TIPA(→)-3B & 61,188 &  $\sim0.58$ s/batch & $\sim10$ h\\
    TIPA(←)-3B & 61,188 &  $\sim0.58$ s/batch & $\sim10$ h\\    
    Pure-SFT-1.5B & 56,502 &  $\sim0.35$ s/batch & $\sim5.5$ h\\    
    TIPA(→)-1.5B & 61,188 &  $\sim0.35$ s/batch & $\sim5.9$ h\\    
    TIPA(←)-1.5B & 61,188 &  $\sim0.35$ s/batch & $\sim5.9$ h\\    
    \bottomrule
  \end{tabular}
  \caption{Training speed and resource consumption for Experiment 2.}
  \label{tab:train_speed_exp2}
\end{table}

\subsubsection{Full-Parameter SFT Configuration}
For general model training without LoRA adaptation, we implemented complete parameter fine-tuning using the configuration shown in Table~\ref{tab:config_full_sft}:

\begin{table}[h]
  \centering
  \small
  \begin{tabular}{lc}
    \toprule
    \textbf{Configs} & \textbf{Values} \\
    \midrule
    Devices & 4 NVIDIA H100 GPUs (80GB) \\
    Batch size & 32 (gradient accumulation) \\
    Learning rate & $5 \times 10^{-6}$ \\
    Epochs & 2 \\ 
    \bottomrule
  \end{tabular}
  \caption{Full-parameter SFT configurations. All unspecified parameters align with TULU training settings. Training completed in 3d 8h 10m 26s.}
  \label{tab:config_full_sft}
\end{table}

\subsection{Datasets}

We utilized several datasets for training and evaluation. The datasets and their characteristics are summarized in Tables~\ref{tab:dataset_train_val} and \ref{tab:dataset_test}. The total number of unique tokens after deduplication across all datasets is 24,994.

\begin{table}[h]
\centering
  \small
\begin{tabular}{lcc}
\toprule
\textbf{Dataset} & \textbf{Samples} & \textbf{Unique Tokens} \\
\midrule
\textbf{Training} & & \\
CSCD-NS Train & 30,000 & 22,012 \\
Wang271K & 271,329 & 20,369  \\
\midrule
\textbf{Validation} & & \\
CSCD-NS Dev & 5,000 & 16,003 \\
\bottomrule
\end{tabular}
\caption{Training and validation datasets were used in our experiments. ``Unique Tokens'' refers to the number of unique tokens appearing in the tokenizer after deduplication.}
\label{tab:dataset_train_val}
\end{table}

\begin{table}[h]
\centering
  \small
\begin{tabular}{lcc}
\toprule
\textbf{Dataset} & \textbf{Filtered Samples} & \textbf{Unique Tokens} \\
\midrule
CSCD-NS Test & 5,000 & 15,946  \\
NOV & 6,000 & 10,503 \\
CAR & 3,245 & 10,881  \\
COT & 993 & 3,564 \\
ENC & 3,274 & 13,399  \\
GAM & 393 & 3,055  \\
MEC & 1,942 & 4,813  \\
NEW & 5,887 & 12,010 \\
\bottomrule
\end{tabular}
\caption{Testing datasets used in our experiments after filtering. ``Filtered Samples'' indicates the number of samples remaining after filtering out those with unequal source and target sentence lengths. ``Unique Tokens'' refers to the number of unique tokens appearing in the tokenizer after deduplication.}
\label{tab:dataset_test}
\end{table}

The distribution of token lengths in the pruned token set used to construct the TIPA dataset is shown in Figure~\ref{fig:token_length_distribution}. We observed that most tokens have lengths of 1 or 2 characters.

\begin{figure}[h]
  \includegraphics[width=\columnwidth]{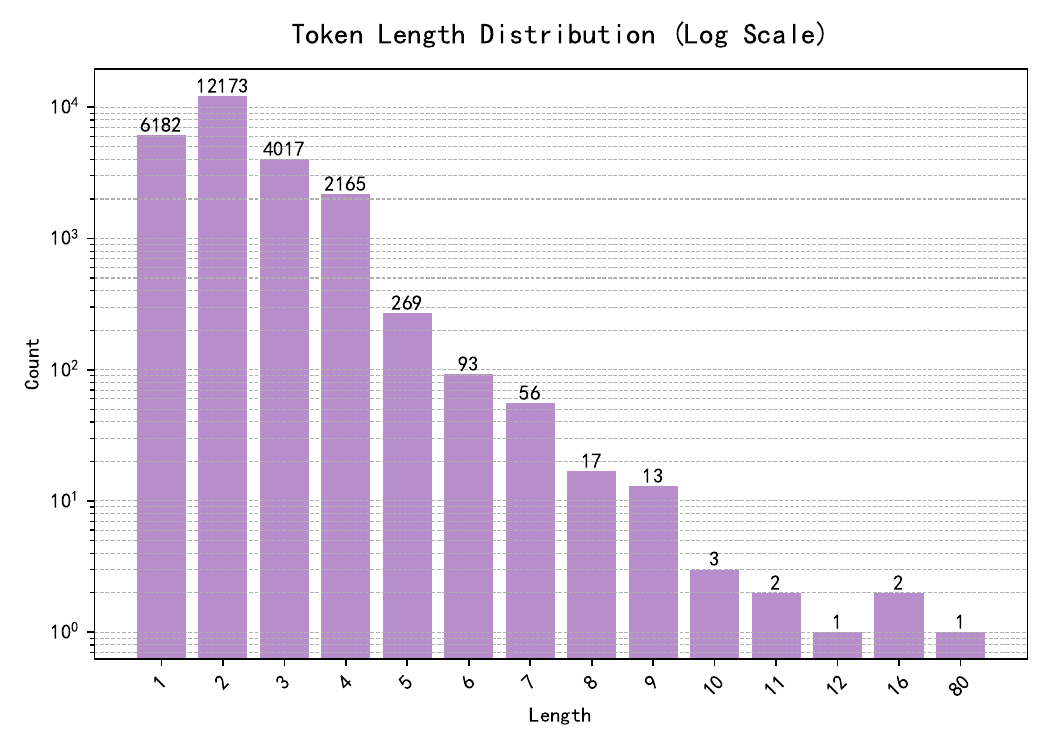}
  \caption{Token length distribution in the pruned token set used for constructing the TIPA dataset. The y-axis is on a logarithmic scale\citep{hunter2007matplotlib}. The token with a length of 80 is an exceptional case and can be excluded in practical applications to avoid excessively long tokens in TIPA.}
  \label{fig:token_length_distribution}
\end{figure}

\subsection{Prompts Used in Experiments}

We provide the prompts used in our experiments for different models and tasks. Due to the length of the prompts, we format them to allow for automatic line wrapping.

\begin{table*}[h]
\centering
\small
\begin{tabular}{p{5cm} p{8cm}}
\toprule
\textbf{Experiment} & \textbf{Prompt (Chinese / English)} (temperature = 0.01) \\
\midrule
\textbf{Experiment 1 (All models)} & 
\textbf{Chinese:} \begin{CJK}{UTF8}{gkai}纠正输入这段话中的错别字，以 \texttt{[\{position: 字符位置, incorrect: 错误字符, correction: 纠正后的字符\}, ...]} 形式给出，字符位置从 1 开始计数，必须是单个字符，如果全部正确，给出 \texttt{[]}。\end{CJK}

\textbf{English:} Correct the typos in this paragraph of input and provide it in the form \texttt{[{position: character position, incorrect: wrong character, correction: corrected character}, ...]}, where character position starts counting from 1, must be a single character, if all is correct, give \texttt{[]}. \\
\midrule
\textbf{Experiment 2 (Testing GPT-4o, Qwen2.5-7B original models) } (temperature = 0.01) & 
\textbf{Chinese:} \begin{CJK}{UTF8}{gkai}纠正输入这段话中的错别字，直接给出纠正后的文本，无需任何解释，不要补充任何标点符号，尽可能输出等长的新句子！\end{CJK}

\textbf{English:} Correct the typos in this paragraph of input and directly give the corrected text, no need for any explanation, do not add any punctuation marks, and try to output a new sentence of equal length! \\
\midrule
\textbf{Experiment 2 (Trained models)} & 
\textbf{Chinese:} \begin{CJK}{UTF8}{gkai}纠正输入这段话中的错别字，直接给出纠正后的文本，无需任何解释。\end{CJK}

\textbf{English:} Correct the typos in this paragraph of input, and directly give the corrected text, no need for any explanation. \\
\bottomrule
\end{tabular}
\caption{Prompt templates used in our experiments, presented in both Chinese and English. The prompts were designed to elicit the desired output formats from the models.}
\label{tab:prompts}
\end{table*}

\subsection{Analysis of Output Length Consistency}

We analyzed the proportion of outputs where the corrected text has the same character length as the original text. This metric reflects the models' ability to maintain length consistency, which is important for certain applications.

\begin{table*}[h]
\centering
\small
\begin{tabular}{lccccccccc}
\toprule
Model & CAR & COT & ENC & GAM & MEC & NEW & NOV & CSCD-NS & AVG \\
\midrule
Qwen2.5-7B & 59.69 & 65.16 & 63.44 & 55.98 & 65.71 & 58.72 & 58.33 & 66.02 & 61.63 \\
GPT-4o & 77.69 & 83.18 & 83.87 & 82.44 & 86.56 & 79.80 & 79.63 & 87.74 & 82.61 \\
\midrule
Pure-SFT-1.5B & \textbf{95.13} & \textbf{96.88} & \textbf{96.06} & 96.18 & \textbf{95.83} & \textbf{97.27} & 96.12 & 96.28 & 96.22 \\
TIPA(→)-1.5B & 95.04 & 95.47 & 95.97 & \textbf{96.95} & 95.67 & 97.06 & \textbf{96.33} & \textbf{98.68} & \textbf{96.40} \\
TIPA(←)-1.5B & 94.67 & 95.57 & 95.97 & 95.42 & 95.62 & 96.96 & 96.0 & 98.52 & 96.09 \\
\midrule
Pure-SFT-3B & 94.82 & 95.37 & 95.66 & 95.67 & 96.55 & 97.50 & 94.97 & \textbf{98.92} & 96.18 \\
TIPA(→)-3B & 95.01 & 95.37 & \textbf{96.46} & 95.93 & 96.14 & \textbf{97.55} & 94.43 & 98.66 & 96.19 \\
TIPA(←)-3B & \textbf{95.50} & \textbf{96.07} & 95.94 & \textbf{97.20} & \textbf{96.86} & 97.40 & \textbf{95.47} & 98.66 & \textbf{96.64} \\
\midrule
Pure-SFT-7B & 93.93 & 95.27 & 95.51 & 96.44 & 93.72 & 97.03 & 94.05 & 98.60 & 95.57 \\
TIPA(→)-7B & \textbf{95.75} & \textbf{96.27} & \textbf{96.30} & \textbf{97.71} & \textbf{96.04} & \textbf{98.05} & \textbf{96.07} & \textbf{99.16} & \textbf{96.92} \\
TIPA(←)-7B & 94.58 & 94.76 & 95.51 & 97.20 & 95.57 & 97.55 & 95.42 & 98.84 & 96.18 \\
\bottomrule
\end{tabular}
\caption{Proportion (\%) of outputs where the corrected text has the same character length as the original text across different domains and models. The average (AVG) is also provided.}
\label{tab:length_consistency}
\end{table*}

In Table~\ref{tab:length_consistency}, Qwen2.5-7B and GPT-4o were explicitly instructed to output texts of the same length as much as possible, but their averages are relatively low, indicating that these models do not fully grasp the length information of the tokens. Our models were not instructed to output texts of the same length, as all training samples were of equal length. This demonstrates the models' ability to mimic and generalize length information through training methods. In CSCD-NS, the error patterns of tokens are present in the training set, so the Pure-SFT-3B model has good mimicking ability. In the LEMON dataset, since more new tokens may appear, TIPA has trained character composition awareness for all possible tokens, while the Pure-SFT method cannot perceive such tokens during training, so TIPA shows better generalization ability. However, this does not involve including the test set in the training set, because the tokens are known in the vocabulary, and we can train the tokenizer's vocabulary without obtaining any datasets.

\subsection{Additional Results}

We provide additional examples from our experiments to illustrate the impact of TIPA and MTIPA on model outputs.

\subsubsection{Experiment 1: CSC with Position Prediction}

Table~\ref{tab:experiment1_examples} shows several examples comparing the outputs of the Pure-SFT-7B model and the MTIPA-7B model on the CSC task with position prediction.

\begin{table*}[h]
\centering
\small
\begin{tabular}{p{0.95\linewidth}}
\toprule
\textbf{Source Text:} \\
\begin{CJK}{UTF8}{gkai}...产品的移动芯片及相关平台开发，\colorbox{false_bg}{\textcolor{white}{期}}芯片类产品支持...\end{CJK} \\
\textbf{Pure-SFT-7B Output:} \\
\begin{CJK}{UTF8}{gkai}[
    \{
        "position": \colorbox{false_bg}{\textcolor{white}{53}},
        "incorrect": "期",
        "correction": "其"
    \}
]\end{CJK} \\
\textbf{MTIPA-7B Output:} \\
\begin{CJK}{UTF8}{gkai}[
    \{
        "position": \colorbox{mycolor}{\textcolor{white}{52}},
        "incorrect": "期",
        "correction": "其"
    \}
]\end{CJK} \\
\textbf{Translation:} "...development of mobile chips and related platforms for products, \colorbox{mycolor}{\textcolor{white}{its}} chip products support..." \\
\textbf{Comment:} The MTIPA-7B model correctly identifies the position of the incorrect character "\begin{CJK}{UTF8}{gkai}期\end{CJK}" (should be "\begin{CJK}{UTF8}{gkai}其\end{CJK}") at position 52, whereas the Pure-SFT-7B model predicts position 53. \\
\midrule
\textbf{Source Text:} \\
\begin{CJK}{UTF8}{gkai}...加盟中超大连实德后\colorbox{false_bg}{\textcolor{white}{再}}球队效力三年，上赛季结束后离队，最终选择退役\end{CJK} \\
\textbf{Pure-SFT-7B Output:} \\
\begin{CJK}{UTF8}{gkai}[
    \{
        "position": \colorbox{false_bg}{\textcolor{white}{38}},
        "incorrect": "再",
        "correction": "在"
    \}
]\end{CJK} \\
\textbf{MTIPA-7B Output:} \\
\begin{CJK}{UTF8}{gkai}[
    \{
        "position": \colorbox{mycolor}{\textcolor{white}{37}},
        "incorrect": "再",
        "correction": "在"
    \}
]\end{CJK} \\
\textbf{Translation:} "...after joining Dalian Shide in the Chinese Super League, he served the team \colorbox{mycolor}{\textcolor{white}{for}} three years, left after last season, and finally chose to retire." \\
\textbf{Comment:} The MTIPA-7B model accurately predicts the position of the incorrect character "\begin{CJK}{UTF8}{gkai}再\end{CJK}" (should be "\begin{CJK}{UTF8}{gkai}在\end{CJK}") at position 37, while the Pure-SFT-7B model predicts position 38. \\
\bottomrule
\end{tabular}
\caption{Examples from Experiment 1 comparing Pure-SFT-7B and MTIPA-7B outputs on the CSC task with position prediction. The MTIPA-7B model demonstrates enhanced position prediction accuracy.}
\label{tab:experiment1_examples}
\end{table*}

\subsubsection{Experiment 2: Traditional CSC Task}

Table~\ref{tab:experiment2_examples} provides examples from Experiment 2, comparing the outputs of the Pure-SFT-7B model and the TIPA-7B model on the traditional CSC task.

\begin{table*}[ht]
\centering
\small
\begin{tabular}{p{0.95\linewidth}}
\toprule
\textbf{Source Tokens:} \\
\begin{CJK}{UTF8}{gkai}.../ 以 / 吃的 / 名 / \colorbox{false_bg}{\textcolor{white}{意}} / 来 / 一场 / 说 / 走 / 就 / 走 / 的 / 美食 / 之旅 / \end{CJK} \\
\textbf{Pure-SFT-7B Output:} \\
\begin{CJK}{UTF8}{gkai}.../ 以 / 吃的 / \colorbox{false_bg}{\textcolor{white}{心情}} / 来 / 一场 / 说 / 走 / 就 / 走 / 的 / 美食 / 之旅 / \end{CJK} \\
\textbf{TIPA-7B Output:} \\
\begin{CJK}{UTF8}{gkai}.../ 以 / 吃的 / \colorbox{mycolor}{\textcolor{white}{名义}} / 来 / 一场 / 说 / 走 / 就 / 走 / 的 / 美食 / 之旅 / \end{CJK} \\
\textbf{Translation:} "Take a spontaneous food journey \colorbox{mycolor}{\textcolor{white}{in the name of}} eating." \\
\textbf{Comment:} The TIPA-7B model correctly reconstructs "\begin{CJK}{UTF8}{gkai}名义\end{CJK}" ("name"), improving the fluency and accuracy of the sentence compared to the Pure-SFT-7B model's output. \\
\midrule
\textbf{Source Tokens:} \\
\begin{CJK}{UTF8}{gkai}/ 伦敦 / 又 / \colorbox{false_bg}{\textcolor{white}{粗}} / \colorbox{false_bg}{\textcolor{white}{线}} / 了一 / 家 / 哈利 / · / 波特 / 主题 / 的 / 酒店 / 套房 / \end{CJK} \\
\textbf{Pure-SFT-7B Output:} \\
\begin{CJK}{UTF8}{gkai}/ 伦敦 / 又 / \colorbox{false_bg}{\textcolor{white}{粗}} / \colorbox{false_bg}{\textcolor{white}{线}} / 了一 / 家 / 哈利 / · / 波特 / 主题 / 的 / 酒店 / 套房 / \end{CJK} \\
\textbf{TIPA-7B Output:} \\
\begin{CJK}{UTF8}{gkai}/ 伦敦 / 又 / \colorbox{mycolor}{\textcolor{white}{出现}} / 了一 / 家 / 哈利 / · / 波特 / 主题 / 的 / 酒店 / 套房 / \end{CJK} \\
\textbf{Translation:} "London has another Harry Potter themed hotel suite \colorbox{mycolor}{\textcolor{white}{appeared}}." \\
\textbf{Comment:} The TIPA-7B model accurately corrects "\begin{CJK}{UTF8}{gkai}粗线\end{CJK}" (nonsensical in this context) to "\begin{CJK}{UTF8}{gkai}出现\end{CJK}" ("appear"), enhancing the sentence's coherence. \\
\midrule
\textbf{Source Tokens:} \\
\begin{CJK}{UTF8}{gkai}/ \colorbox{false_bg}{\textcolor{white}{本事}} / 一家人 / ...\end{CJK} \\
\textbf{Pure-SFT-7B Output:} \\
\begin{CJK}{UTF8}{gkai}/ 本 / 是 / \colorbox{false_bg}{\textcolor{white}{亲}} / 一家 / ... \end{CJK} \\
\textbf{TIPA-7B Output:} \\
\begin{CJK}{UTF8}{gkai}/ 本 / \colorbox{mycolor}{\textcolor{white}{是一}} / 家人 / ...\end{CJK} \\
\textbf{Translation:} "We \colorbox{mycolor}{\textcolor{white}{were}} originally a family." \\
\textbf{Comment:} The TIPA-7B model correctly reconstructs the phrase "\begin{CJK}{UTF8}{gkai}是一家人\end{CJK}" ("are a family"), improving the grammaticality and meaning of the sentence. \\
\bottomrule
\end{tabular}
\caption{Examples from Experiment 2 comparing Pure-SFT-7B and TIPA-7B outputs on the traditional CSC task. The TIPA-7B model demonstrates an enhanced understanding of token internal structures, leading to better corrections.}
\label{tab:experiment2_examples}
\end{table*}

\subsection{Loss Analysis}

Figure~\ref{fig:combined_loss_plots} compares training and validation loss across different methods. The TIPA-7B model exhibits the fastest loss reduction during training, indicating more efficient learning.

\begin{figure}[h]
  \includegraphics[width=\columnwidth]{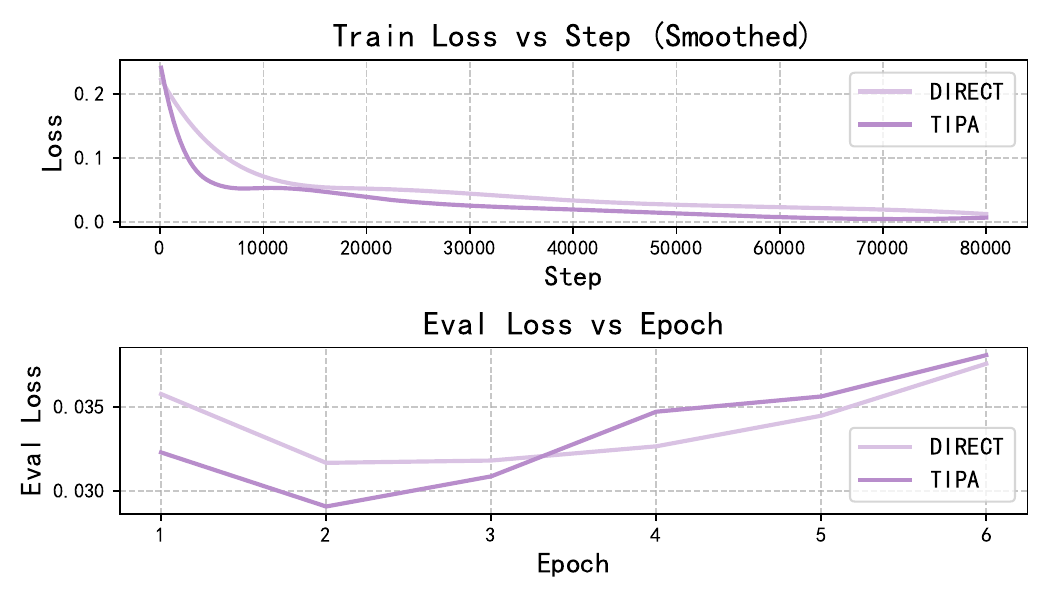}
  \caption{(Experiment 2)Training and validation loss comparison across different methods in the traditional CSC task. The TIPA-7B model exhibits the fastest reduction in loss during training, indicating more efficient learning and better generalization capabilities compared to the baseline.}
  \label{fig:combined_loss_plots}
\end{figure}

\end{document}